\documentclass[runningheads]{llncs}
\usepackage[T1]{fontenc}
\usepackage{amsmath}
\usepackage{subcaption}
\usepackage{svg}
\usepackage{amssymb}
\usepackage{graphicx}
\usepackage[outdir=./]{epstopdf}
\usepackage{float}

\begin{document}

%Semantically Stable Image Composition Analysis via Saliency and Vector Field Fusion\thanks{Supported by organization x.}
\title{Semantically Stable Image Composition Analysis via Saliency and Gradient Vector Flow Fusion}
\author{Armin Dadras\inst{1,2}\orcidID{0000-0001-6474-7208} \and
Robert Sablatnig\inst{1}\orcidID{0000-0003-4195-1593} \and Franziska Proksa\inst{2}\orcidID{0000-0003-3455-8582} \and 
Markus Seidl\inst{2}\orcidID{0000-0002-7966-3602}}
\authorrunning{F. Author et al.}
\institute{Computer Vision Lab, TU Wien \\
Vienna, Austria\\
\email{\{adadras, sab\}@cvl.tuwien.ac.at}\\
\and
UAS St. Pölten\\ St. Pölten, Austria\\
\email{\{armin.dadras,franziska.proska, markus.seidl\}@ustp.at}}
\maketitle              % typeset the header of the contribution
\begin{abstract}
The reliable computational assessment of photographic composition requires features that are discriminative of spatial layout yet robust to semantic content. This paper proposes a low-level representation grounded in the assumption that composition can be understood as the flow of visual attention across geometric structure. We introduce VFCNet, which fuses saliency and edge information into a gradient vector flow (GVF) field. The model computes dual-stream GVF representations, integrates them via attention, and extracts multi-scale flow features with a DINOv3 backbone. VFCNet achieves state-of-the-art performance on the PICD benchmark (CDA-1: 0.683, CDA-2: 0.629), improving by 33.1\% and 36.1\% over the previous best method. We also show that a simple classifier on self-supervised DINOv3 features substantially outperforms more sophisticated, composition-specialized models. Code is available at \url{https://github.com/ADadras/VFCNet}.
\end{abstract}
\section{Introduction}
The reliable computational assessment of image composition remains a significant challenge in computer vision. This task aims to quantify established photographic principles, such as the rule of thirds, vanishing points, and geometric balance. It has practical applications in automatic photo editing \cite{hong2021composing}, image retrieval \cite{zhao2024self}, or aesthetic evaluation \cite{She_2021_CVPR}. A recent benchmark with the name Photographic Image Composition Dataset (PICD) has revealed critical limitations in existing methodologies \cite{zhao2025can}. Evaluations based on Composition Discrimination Accuracy (CDA) demonstrate that models achieving high accuracy on smaller, dedicated classification datasets often perform close to random chance when evaluated for robust composition embedding, particularly when semantic interference is introduced.

The analysis of the PICD benchmark offers several insights into effective composition modelling. First, the top-performing models are not necessarily trained for direct composition classification but often learn compositional cues as part of auxiliary tasks (e.g., aesthetic assessment or image cropping). Second, these models frequently employ architectures that capture hierarchical or multi-scale features. A third, more general consideration is the conceptual separation between low-level structural patterns and high-level semantic content, suggesting these features may require distinct processing pathways.

This work investigates a novel approach guided by these observations. Similar to the authors of the PICD paper we take inspiration from Kandinskys concept of composition, where an image can be understood as a dynamic flow of forces \cite{kandinsky1947point}. First, we deliberately avoid overfitting to a specific training dataset by employing a simple classifier head with fixed, general-purpose hyperparameters. Second, we explicitly design our model to fuse multi-level visual features extracted from a pre-trained backbone. Third, and most centrally, we explicitly separate structural and content-related information by operating on a derived input modality. Our central hypothesis is that this saliency- and edge-driven representation contains sufficient information to achieve state-of-the-art performance on the CDA benchmark. Our contributions are threefold:

\begin{itemize}
    \item \textbf{VFCNet architecture:} We introduce the Vector Flow Composition Network, combining a vision transformer backbone with dual-stream gradient flow extraction. VFCNet achieves new SOTA on PICD, improving CDA-1 by 33.1\% and CDA-2 by 36.1\% over previous best (CACNet).
    
    \item \textbf{DINOv3 baseline:} We demonstrate that self-supervised DINOv3 features inherently encode composition. Even DINOv3 without fine-tuning matches specialized models; fine-tuned (DINOv3+C) substantially outperforms them.
    
    \item \textbf{Comprehensive validation:} We prove the effectiveness of our approach through ablations on input modalities, GVF parameters, edge sources, saliency models, and architecture components, demonstrating that edges and saliency are sufficient for composition understanding.
\end{itemize}
\section{Related Work}
Features related to the visual composition of an image are typically learned through two primary approaches: direct supervision on composition classification, or indirect supervision via auxiliary tasks that inherently require compositional understanding.
\subsection{Models trained for Composition}
The computational analysis of photographic composition classifies images based on rules governing the spatial arrangement of visual elements. The field progresses with the creation of annotated datasets and corresponding deep learning models. The first large-scale benchmark, the KU-PCP dataset, enables the training of deep networks for this task \cite{lee2018photographic}. The established baseline method involves fine-tuning a CNN pre-trained on ImageNet for multi-label classification on KU-PCP \cite{lee2018photographic}.
Subsequent research develops architectures that integrate auxiliary tasks. CACNet introduces a dual-branch network performing joint composition classification and aesthetic cropping, guided by a learned Key Composition Map \cite{hong2021composing}. The Synchronous Detection Network employs a multi-task learning framework with cross-attention to synchronize composition classification with semantic line detection, utilizing knowledge distillation from pre-trained teachers \cite{hou2024synchronous}. To improve robustness against spatial perturbations such as rotation, the Spatial-Invariant CNN incorporates a Rotation-Shift Transformer Network module to align features \cite{wang2023spatial}.
Shifting focus from classification to quality assessment, the CADB dataset provides composition scores instead of class labels \cite{zhang2021image}. This regression task is addressed by SAMP-Net using a Saliency-Augmented Multi-pattern Pooling module with a weighted Earth Mover's Distance loss. Subsequent work introduces the LODB dataset, offering finer-grained layout-oriented classes \cite{zhao2024self}. The associated method represents images as heterogeneous graphs of structural and object primitives, learning layout embeddings for retrieval via self-supervised graph autoencoders. PICD is a dataset introduced to test the generalizability of existing methods by offering more data and classes \cite{zhao2025can}. It includes a custom metric that relies on retrieval and enables comparing the features of specialised models and VLMs. 
\subsection{Aesthetic, Image‑Quality and Cropping Models}
Methods evaluated on composition also originate from aesthetics assessment and image cropping.  The Hierarchical Layout‑Aware Graph Convolutional Network (HLGCN)~\cite{She_2021_CVPR} treats an image as a graph of detected objects and applies hierarchical attention to predict aesthetic scores using the AVA and AADB datasets.  MUSIQ~\cite{Ke_2021_ICCV} leverages a transformer to compute multi‑scale representations; it is pre‑trained on ImageNet and fine‑tuned on several quality datasets (PaQ‑2‑PiQ, SPAQ, KonIQ‑10k and AVA).  Yi et al. introduce the Style‑specific Art Assessment Network (SAAN)~\cite{Yi_2023_CVPR}, training it on the Boldbrush Artistic Image Dataset (BAID, 60k artworks) alongside other aesthetic corpora.  EAT~\cite{DBLP:conf/mm/HeMZZM23} adds deformable sparse attention to improve transformer‑based aesthetic assessment; training uses the AVA, TAD66K, FLICKR‑AES and AADB datasets.

Cropping models learn to select pleasing sub‑windows rather than classify compositions directly.  CGS~\cite{Li_2020_CVPR} employs a graph neural network to model relations among candidate crops and is trained on GANCD, which provides dense annotations for 1\,036 images.  GANC~\cite{Zeng_2022_TPAMI} formulates cropping as predicting scores over a dense grid of anchors; the training set comprises roughly 1\,236 images with 106\,860 annotated crops.  S2CNet~\cite{su2024spatialsemanticcollaborativecroppinguser} reasons over spatial and semantic relations between candidate crops via a graph module; it uses the UGCrop5K dataset, containing 5\,000 images and 450\,000 crop annotations drawn from varied user‑generated content.
\section{Method}
\label{sec:method}
\begin{figure}
    \centering
    \includegraphics[width=1.1\linewidth]{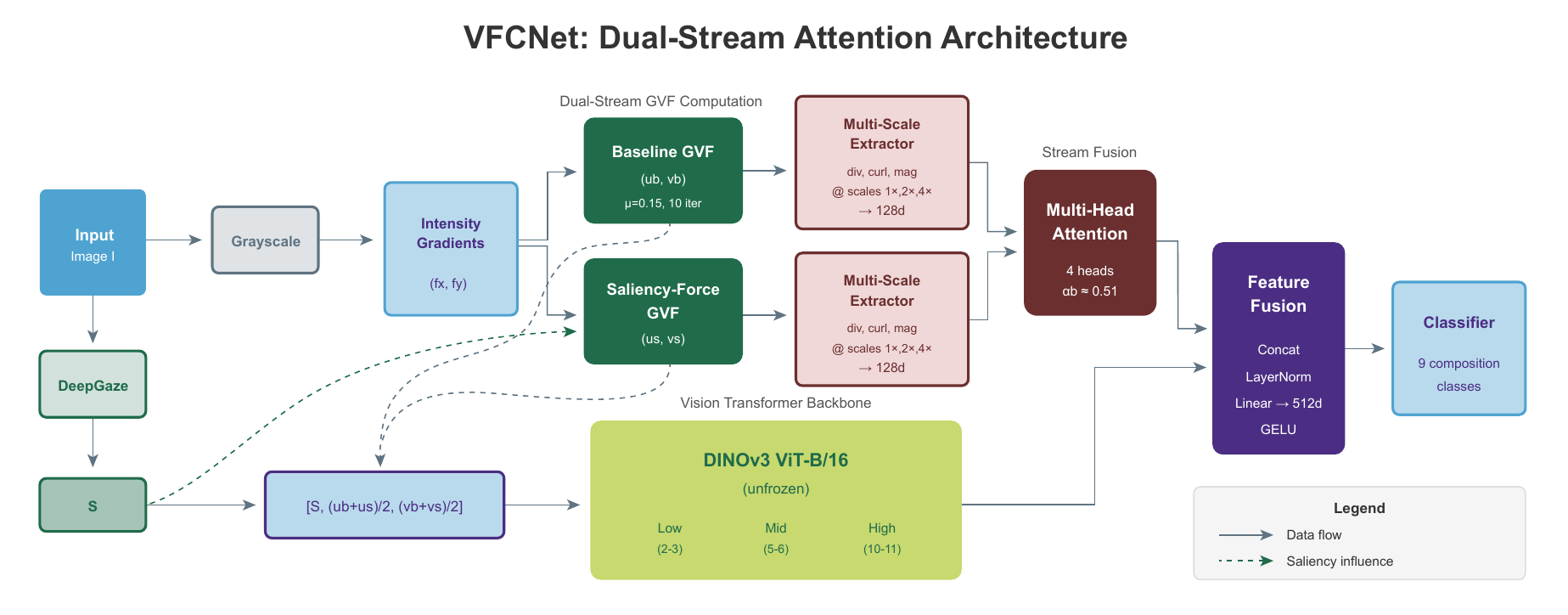}
    \caption{Overview of VFCNet Architecture: saliency and gradient based GVF are fused in the multi-head module and serve as input to the DINOv3 feature extractor. }
    \label{fig:arch}
\end{figure}
We propose VFCNet (Vector Flow Composition Network), depicted in figure \ref{fig:arch}, a framework that represents image composition as the interplay of visual saliency and edge-based gradient vector flow (GVF). Our approach computes two complementary GVF representations---baseline and saliency-enhanced---and fuses them via learned attention, capturing both structural flow patterns and saliency-guided compositional cues.
\subsection{Overview}
\begin{figure}[t]
    \centering
    \begin{subfigure}{0.23\linewidth}
        \includegraphics[width=\linewidth]{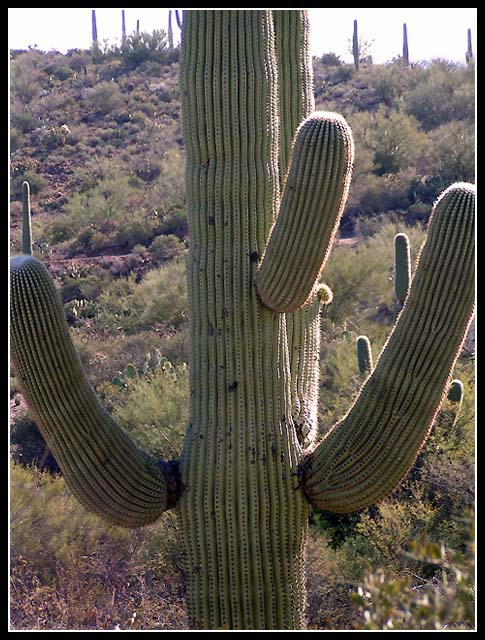}
        \caption{Original}
    \end{subfigure}
    \hfill
    \begin{subfigure}{0.23\linewidth}
        \includegraphics[width=\linewidth]{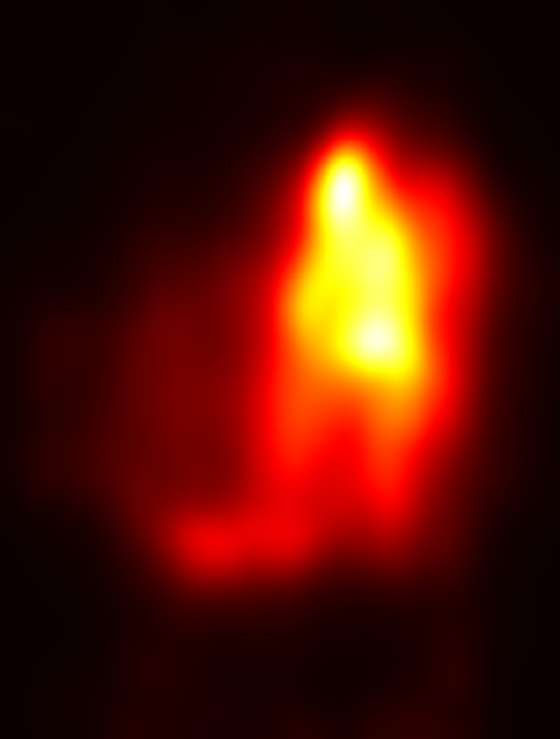}
        \caption{Saliency}
    \end{subfigure}
    \hfill
    \begin{subfigure}{0.23\linewidth}
        \includegraphics[width=\linewidth]{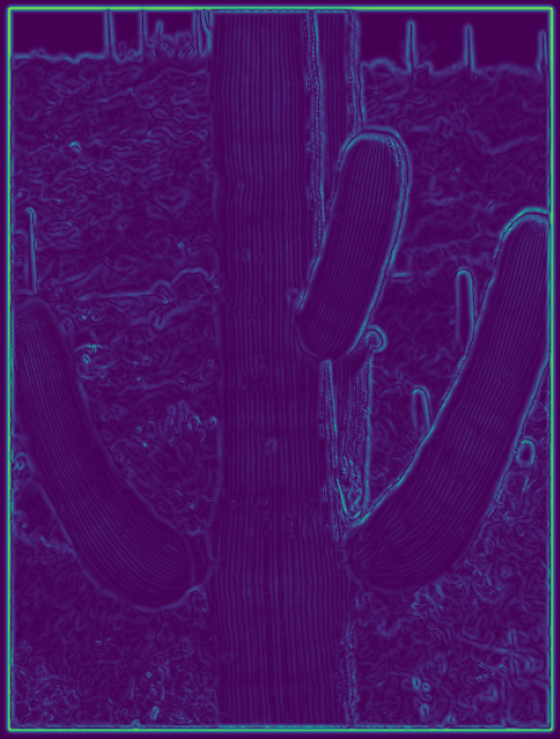}
        \caption{GVF Baseline}
    \end{subfigure}
    \hfill
    \begin{subfigure}{0.23\linewidth}
        \includegraphics[width=\linewidth]{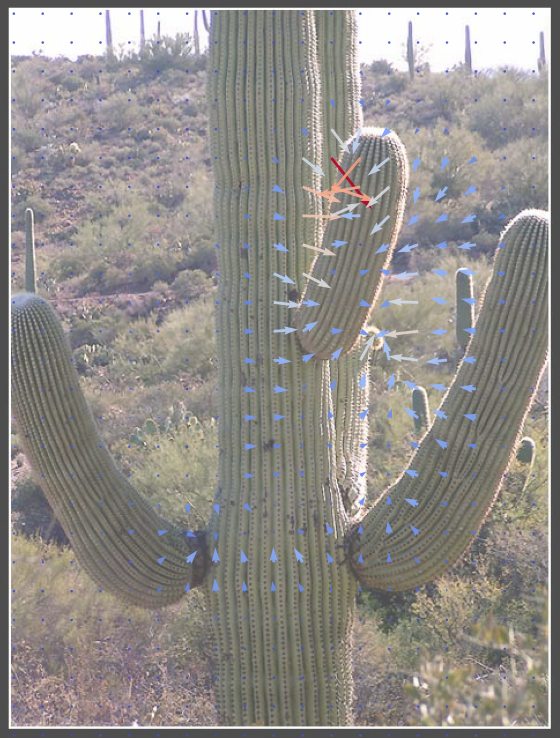}
        \caption{Difference}
    \end{subfigure}
    \caption{VFCNet input representation: saliency captures attention distribution; GVF propagates edge structure throughout the image. The prominent feature of cactus draws attention as visible in (b) and modifies the flow as visible in (d).}
    \label{fig:input_modality}
\end{figure}
As shown in Fig.~\ref{fig:input_modality}, given an input image $I \in \mathbb{R}^{H \times W \times 3}$, our method: (1) computes saliency map $S$ using DeepGaze~\cite{linardos2021deepgaze}, (2) derives dual-stream GVF from intensity gradients, (3) fuses streams via multi-head attention, and (4) extracts multi-scale features combining DINOv3~\cite{simeoni2025dinov3} with explicit flow analysis.
\subsection{Saliency Map Computation}
We compute a saliency map $S \in \mathbb{R}^{H \times W}$ using DeepGaze IIE~\cite{linardos2021deepgaze}, encoding human fixation likelihood:
\begin{equation}
    S = \text{DeepGaze}(I), \quad S \in [0,1]^{H \times W}
\end{equation}
\subsection{Gradient Vector Flow Formulation}
GVF extends edge information throughout the image by minimizing an energy functional that balances smoothness against fidelity to the original gradient field. We first convert to grayscale $I_g$ and compute spatial gradients:
\begin{equation}
    f_x = \frac{\partial I_g}{\partial x}, \quad f_y = \frac{\partial I_g}{\partial y}
\end{equation}
\subsubsection{Baseline GVF Stream}
The baseline GVF $(u_b, v_b)$ minimizes the energy functional:
\begin{equation}
    \mathcal{E}_b = \iint \mu(u_{xx}^2 + u_{yy}^2 + v_{xx}^2 + v_{yy}^2) + |\nabla I_g|^2 |(\mathbf{v} - \nabla I_g)|^2 \, dx\, dy
\end{equation}
where the first term encourages smoothness and the second enforces fidelity to the original gradient where $|\nabla I_g|$ is large. We solve via iterative diffusion:
\begin{equation}
    u_b^{t+1} = u_b^t + \mu \nabla^2 u_b^t - (f_x^2 + f_y^2)(u_b^t - f_x)
    \label{eq:gvf_baseline_u}
\end{equation}
\begin{equation}
    v_b^{t+1} = v_b^t + \mu \nabla^2 v_b^t - (f_x^2 + f_y^2)(v_b^t - f_y)
    \label{eq:gvf_baseline_v}
\end{equation}
where $\nabla^2$ is the Laplacian operator. This stream captures pure structural flow patterns from image edges.
\subsubsection{Saliency-Enhanced GVF Stream}
The saliency-enhanced GVF $(u_s, v_s)$ incorporates saliency gradient as an additional force term, biasing flow toward salient regions. We first compute the saliency gradient:
\begin{equation}
    \nabla S = \left(\frac{\partial S}{\partial x}, \frac{\partial S}{\partial y}\right) = (S_x, S_y)
\end{equation}

This modifies the energy functional to include a saliency attraction term:
\begin{equation}
    \mathcal{E}_s = \mathcal{E}_b - \beta \iint \mathbf{v} \cdot \nabla S \, dx\, dy
\end{equation}
The corresponding update rules become:
\begin{equation}
    u_s^{t+1} = u_s^t + \mu \nabla^2 u_s^t - (f_x^2 + f_y^2)(u_s^t - f_x) + \beta \frac{\partial S}{\partial x}
    \label{eq:gvf_saliency_u}
\end{equation}
\begin{equation}
    v_s^{t+1} = v_s^t + \mu \nabla^2 v_s^t - (f_x^2 + f_y^2)(v_s^t - f_y) + \beta \frac{\partial S}{\partial y}
    \label{eq:gvf_saliency_v}
\end{equation}
where $\beta = 0.1$ controls the saliency force strength. The saliency gradient term acts as a directional force pulling flow vectors toward regions of increasing saliency, encoding compositional focus.
\subsubsection{Parameterization}
Both streams use $\mu = 0.15$ (smoothness weight) and 10 iterations. We use intensity gradients rather than edge maps (Canny/Sobel), as ablations show superior performance at this iteration count (Sec.~\ref{subsec:ablations}). All fields are normalized to $[0, 1]$.
\subsection{Input Representation}
We construct a 3-channel input tensor by averaging both GVF streams:
\begin{equation}
    X = \left[S, \frac{u_b + u_s}{2}, \frac{v_b + v_s}{2}\right] \in \mathbb{R}^{3 \times 224 \times 224}
\end{equation}
This provides the vision transformer with both attention distribution and structural flow.
\subsection{Architecture}
\begin{figure}[t]
    \centering
    \begin{subfigure}{0.45\linewidth}
        \includegraphics[width=\linewidth]{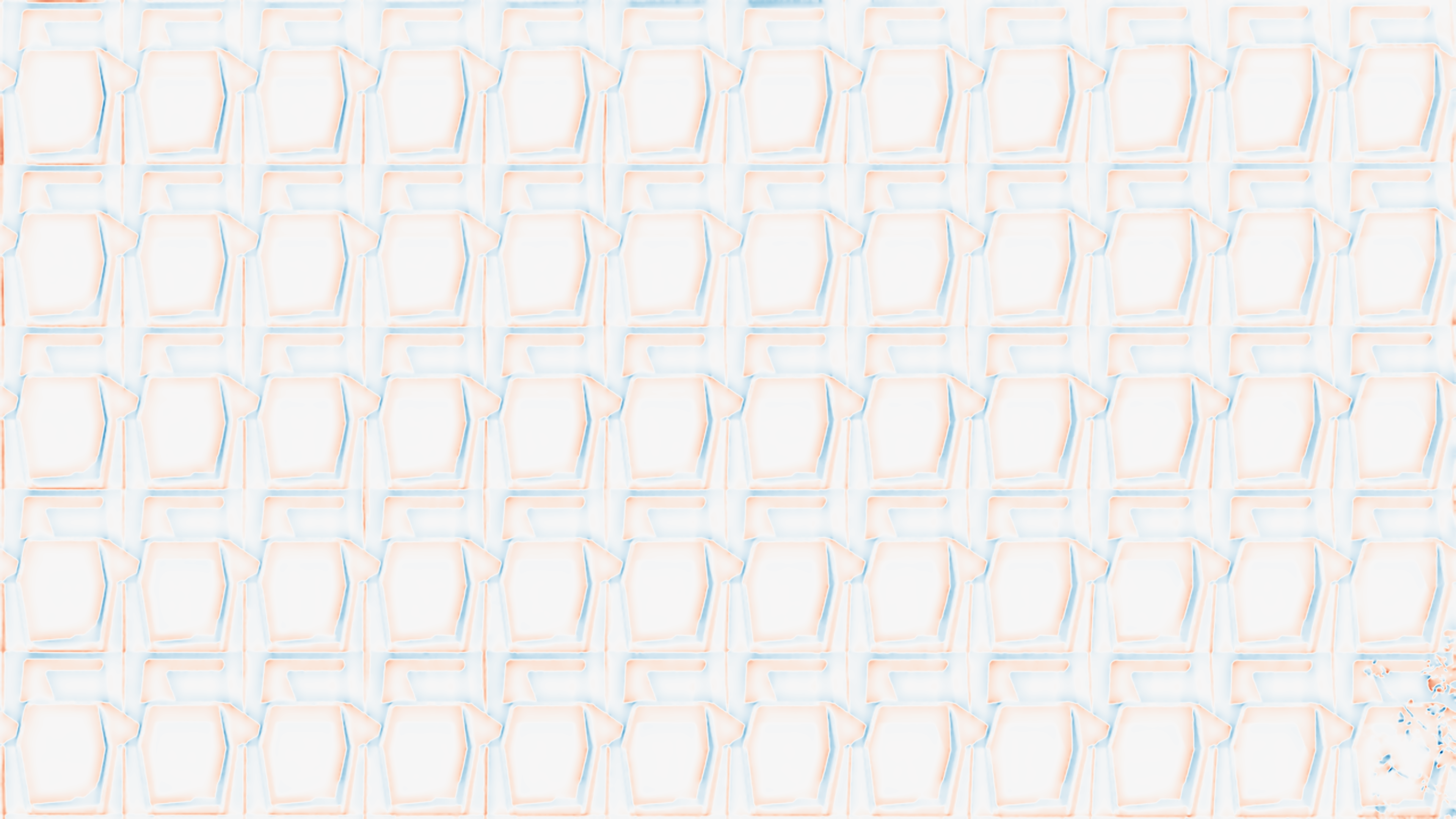}
        \caption{Divergence}
    \end{subfigure}
    \hfill
    \begin{subfigure}{0.45\linewidth}
        \includegraphics[width=\linewidth]{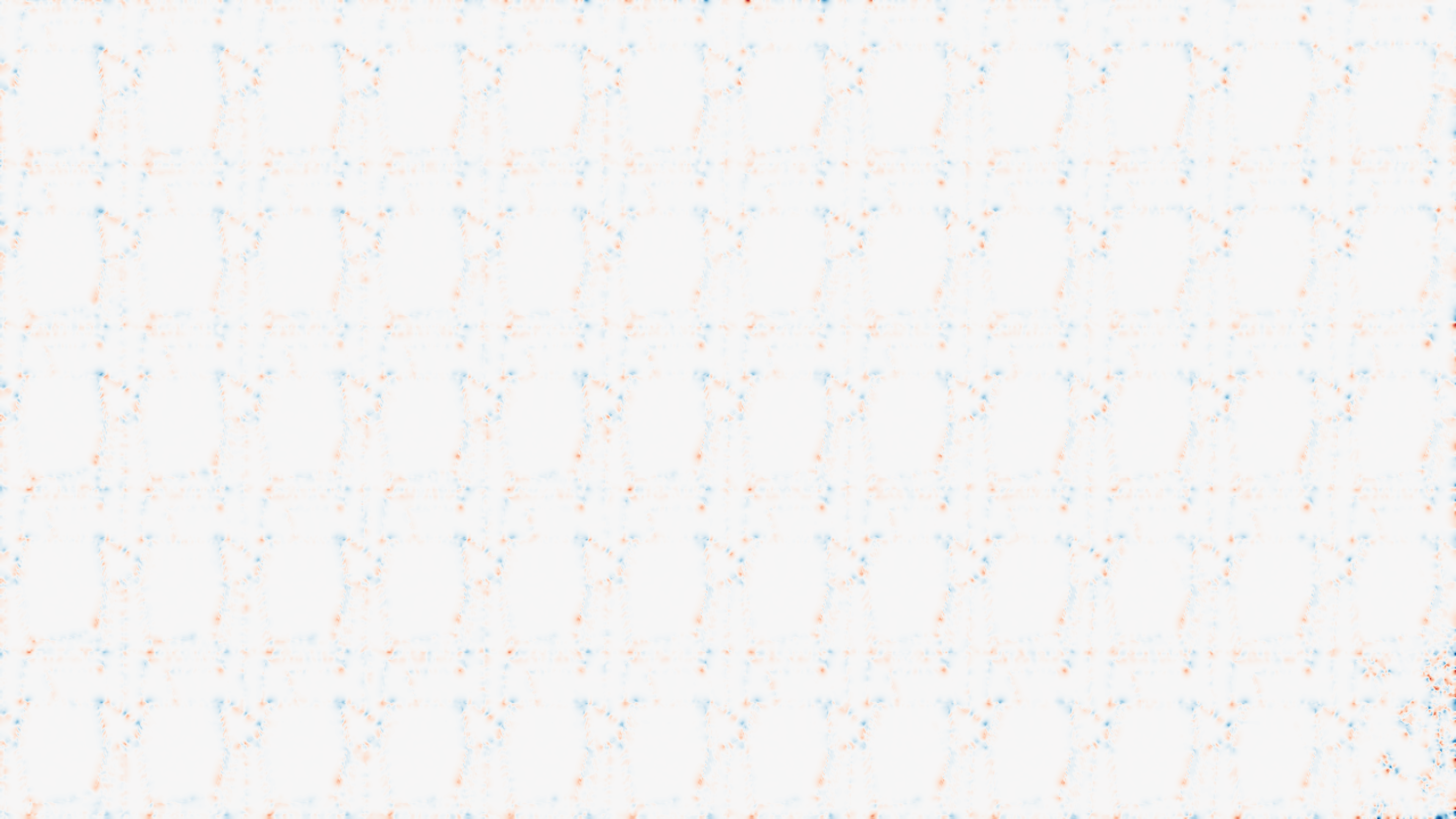}
        \caption{Curl}
    \end{subfigure}
    \hfill
    \begin{subfigure}{0.45\linewidth}
        \includegraphics[width=\linewidth]{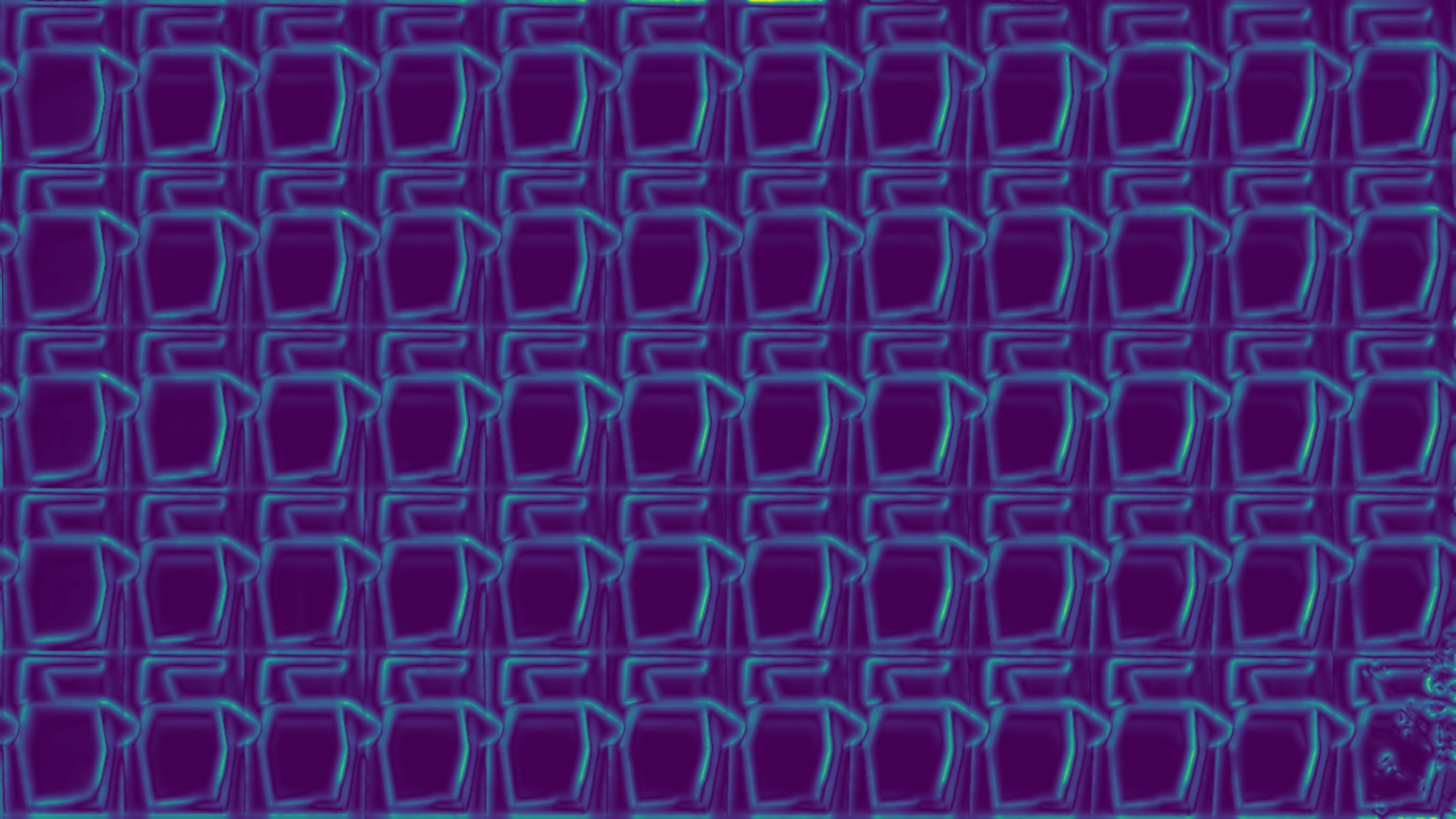}
        \caption{Magnitude}
    \end{subfigure}
    \hfill
    \begin{subfigure}{0.45\linewidth}
        \includegraphics[width=\linewidth]{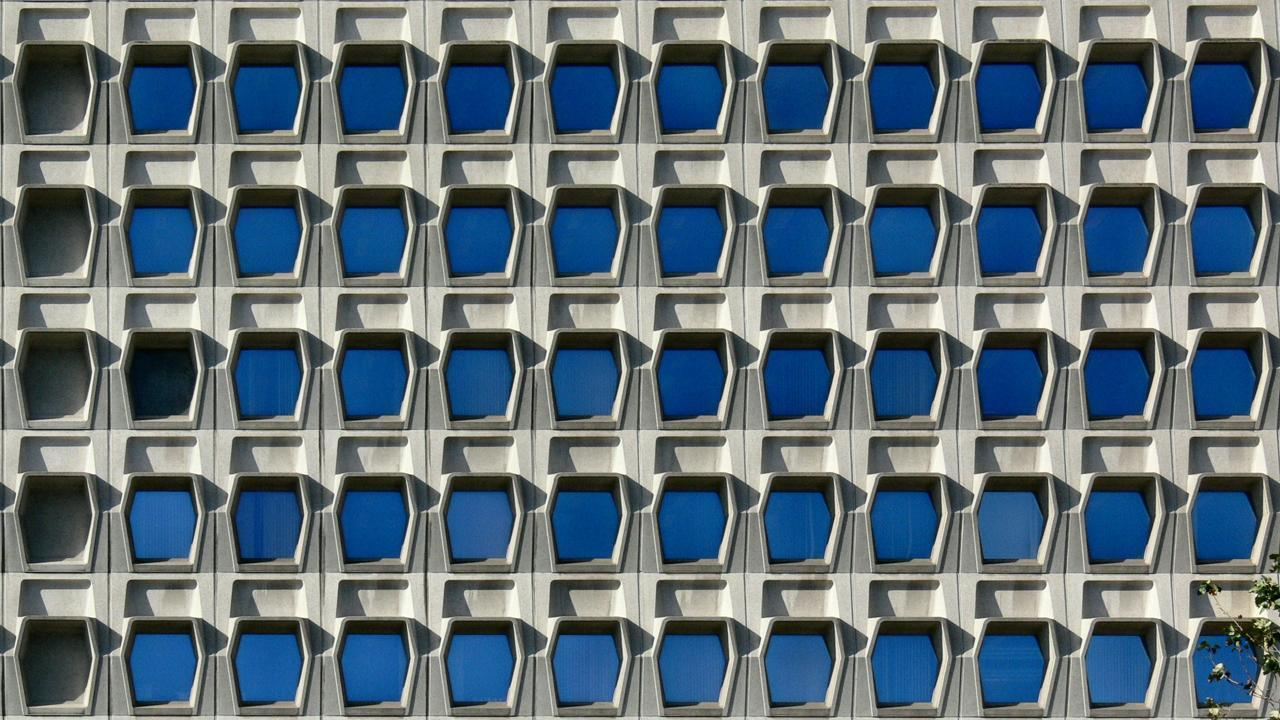}
        \caption{Original}
    \end{subfigure}
    \caption{Multi-scale GVF features: divergence captures convergence/divergence of flow, curl captures rotational patterns, magnitude captures flow strength.}
    \label{fig:gvf_features}
\end{figure}
\subsubsection{Vision Transformer Backbone}
We employ DINOv3 ViT-B/16 with multi-level feature extraction from blocks 2-3 (low-level edges), 5-6 (mid-level structure), and 10-11 (high-level relationships). For each level $l$, we apply global average pooling over spatial tokens:
\begin{equation}
    f_l = \text{Pool}(\text{Blocks}_l(X)) \in \mathbb{R}^{768}
\end{equation}
producing $f_{\text{backbone}} = [f_{\text{low}}; f_{\text{mid}}; f_{\text{high}}] \in \mathbb{R}^{2304}$. We fine-tune the backbone to bridge the RGB$\to$GVF domain gap (Sec.~\ref{subsec:ablations}).
\subsubsection{Dual-Stream GVF Extractor}
Each GVF stream is processed at 56$\times$56 resolution across three scales $s \in \{1, 2, 4\}$ (via average pooling). At each scale, we compute differential features:
\begin{equation}
    \text{div}_s = \frac{\partial u}{\partial x} + \frac{\partial v}{\partial y}, \quad 
    \text{curl}_s = \frac{\partial v}{\partial x} - \frac{\partial u}{\partial y}, \quad 
    \text{mag}_s = \sqrt{u^2 + v^2}
\end{equation}
Examples of these multi-scale GVF descriptors are shown in Fig.~\ref{fig:gvf_features}.
These features are processed by small CNNs for each stream $k \in \{b, s\}$:
\begin{equation}
    g_s^{(k)} = \text{Conv}_{32}(\text{Conv}_{32}([\text{div}_s^{(k)}, \text{curl}_s^{(k)}, \text{mag}_s^{(k)}]))
\end{equation}
Multi-scale features are concatenated with 12 statistical summaries (mean, std, positive/negative ratios per field) and projected:
\begin{equation}
    f_{\text{gvf}}^{(k)} = \text{Linear}([g_1^{(k)}; g_2^{(k)}; g_4^{(k)}; \text{stats}^{(k)}]) \in \mathbb{R}^{128}
\end{equation}
\subsubsection{Attention-Based Stream Fusion}
The two stream features $f_{\text{gvf}}^{(b)}, f_{\text{gvf}}^{(s)}$ are fused via 4-head attention with learnable query $q \in \mathbb{R}^{128}$:
\begin{equation}
    K = V = [f_{\text{gvf}}^{(b)}; f_{\text{gvf}}^{(s)}] \in \mathbb{R}^{2 \times 128}, \quad
    f_{\text{gvf}} = \text{MultiHeadAttn}(q, K, V)
\end{equation}
Empirically, learned weights are approximately balanced ($\alpha_b \approx 0.51$, $\alpha_s \approx 0.49$), suggesting complementary contributions.
\subsubsection{Classification}
Features are concatenated and classified:
\begin{equation}
    f = [f_{\text{backbone}}; f_{\text{gvf}}] \in \mathbb{R}^{2432}
\end{equation}
\begin{equation}
    z = \text{Dropout}(\text{GELU}(\text{Linear}(\text{LayerNorm}(f)))) \in \mathbb{R}^{512}
\end{equation}
\begin{equation}
    \hat{y} = \sigma(\text{MLP}(z)) \in [0,1]^9
\end{equation}
where $\sigma$ is sigmoid for multi-label prediction over 9 composition categories.
\subsection{Training}
\subsubsection{Loss Function}
We train with binary cross-entropy loss for multi-label composition classification:
\begin{equation}
    \mathcal{L} = -\frac{1}{NC} \sum_{i=1}^{N} \sum_{c=1}^{C} \left[ y_{ic} \log(\hat{y}_{ic}) + (1 - y_{ic}) \log(1 - \hat{y}_{ic}) \right]
\end{equation}
\subsubsection{Optimization}
We use AdamW (weight decay 0.01) with differential learning rates: $5\times10^{-5}$ for the classification head and $5\times10^{-6}$ for the backbone (0.1$\times$). We apply cosine annealing over 30 epochs, batch size 16, and gradient clipping (max norm 1.0).
\subsubsection{Data Preprocessing}
Images are resized to $224\times224$ using bilinear interpolation. For training, we apply random horizontal flip (p=0.5) and random crop ($[0.8, 1.0]$ scale). GVF fields are computed at $56\times56$ resolution. All inputs (saliency, GVF) are normalized using ImageNet statistics. The same preprocessing is used for all baselines to ensure fair comparison.
\subsection{RGB Baseline}
For comparison, we train DINOv3+C: the same DINOv3 backbone processing RGB images with a simple MLP head (2304$\to$512$\to$256$\to$9). This isolates the contribution of our saliency-GVF representation versus semantic RGB features.
% VFCNet: Visual Flow Composition Network
\section{Experiments and Results}
\label{sec:experiments}
We evaluate VFCNet following the PICD benchmark protocol and conduct extensive ablations to validate its design. 
\subsection{Experimental Setup}
\label{subsec:setup}
\subsubsection{Datasets}
\textbf{Training:} We train VFCNet on KUPCP~\cite{lee2018photographic}, containing 4,244 images with 9 composition categories (Rule-of-Thirds, Center, Horizontal, Symmetric, Diagonal, Curved, Vertical, Triangle, Pattern). We use the standard 70\%/20\%/10\% train/val/test split.

\textbf{Evaluation:} We evaluate on PICD~\cite{zhao2025can}, a large-scale composition benchmark.  The original PICD paper reports 36,857 images with 24 composition categories and curated object annotations. For our experiments, we use an updated version from the official repository containing 43,444 images, which does not include object-level labels. To enable CDA-2 evaluation (which requires semantic labels for semantic interference), we automatically annotated this set using YOLOv8 \cite{yaseen2024yolov9}. Our YOLO labels cover 79 object categories (e.g., person, car, dog), with 64.4\% of images having at least one detection. For reproducibility, we provide the mapping between image IDs and our generated labels.
\subsubsection{Evaluation Protocol}
We use Composition Distance Assessment (CDA)~\cite{zhao2025can}, measuring whether learned features respect compositional similarity via triplets $(a, p, n)$ where anchor $a$ and positive $p$ share composition class while negative $n$ differs:
\begin{equation}
    \text{CDA-1} = \mathbb{E}_{(a,p,n)} \left[ \mathbf{1}[d(f_a, f_p) < d(f_a, f_n)] \right]
\end{equation}
\textbf{CDA-2} adds semantic interference---anchor and negative share semantic labels---testing composition independence from scene content. 

\textbf{Triplet Sampling:} Triplets are sampled randomly with 12 triplets per image pair. Distance $d$ is computed using L2 (Euclidean) norm. We report mean $\pm$ std over 5 random seeds (42--46), with coefficient of variation (CV) below 3\% for all models. Table~\ref{tab:embeddings} documents exact embeddings for reproducibility.
\begin{table}[t]
\centering
\caption{Embedding specification for CDA evaluation.}
\label{tab:embeddings}
\begin{tabular}{|l|c|c|}
\hline
\textbf{Model} & \textbf{Layer} & \textbf{Dim} \\
\hline
DINOv3 ViT-B/16 & CLS token (block 11) & 768 \\
VFCNet & After dual-stream fusion & 512 \\
DINOv3+C / RGB baselines & After MLP head & 512 \\
\hline
\end{tabular}
\end{table}
\subsubsection{Baseline Models}
We compare against the three best networks from the original PICD paper: \textbf{MUSIQ}~\cite{Ke_2021_ICCV} (multi-scale quality assessment), \textbf{GANC}~\cite{Zeng_2022_TPAMI} (grid anchor cropping), \textbf{CACNet}~\cite{hong2021composing} (composition-aware cropping). We additionally consider stand-alone \textbf{DINOv3}~\cite{simeoni2025dinov3} (self-supervised ViT-B/16). \textbf{DINOv3+C} denotes DINOv3 fine-tuned with a composition classifier. All models are evaluated on our version of the dataset. 
%==============================================================================
\subsection{Main Results}
\label{subsec:main_results}
\begin{table}[h]
\caption{SOTA comparison. CDA-2 is primary metric (semantic robustness). VFCNet achieves best performance on both metrics. Results are mean$\pm$std over 5 seeds. (P) refers to the value in the original evaluation. }
\label{tab:sota_comparison}
\centering
\begin{tabular}{|l|c|c|c|c|c|c|}
\hline
\textbf{Model} & \textbf{CDA-1} & \textbf{CDA-2} & \textbf{DBI$\downarrow$} & \textbf{Sil.$\uparrow$} & \textbf{CDA-1 (P)} & \textbf{CDA-2 (P)} \\
\hline
DINOv3 (base) & $0.489{\pm}.002$ & $0.392{\pm}.008$ & 12.34 & 0.004 & -- & -- \\
MUSIQ & $0.434{\pm}.002$ & $0.369{\pm}.007$ & 12.09 & $-$0.079 & 0.488 & 0.367 \\
GANC & $0.456{\pm}.005$ & $0.423{\pm}.015$ & 9.64 & $-$0.154 & 0.478 & 0.445 \\
CACNet & $0.513{\pm}.005$ & $0.462{\pm}.006$ & 7.36 & $-$0.156 & 0.419 & 0.405 \\
\hline
DINOv3+C & $0.621{\pm}.003$ & $0.505{\pm}.003$ & 5.10 & $-$0.152 & -- & -- \\
\textbf{VFCNet (Ours)} & $\mathbf{0.683{\pm}.002}$ & $\mathbf{0.629{\pm}.004}$ & \textbf{4.01} & $-$0.664 & -- & -- \\
\hline
\end{tabular}
\end{table}
Table~\ref{tab:sota_comparison} presents quantitative comparisons. VFCNet achieves CDA-1=0.683 and CDA-2=0.629, improving over CACNet (previous SOTA) by 33.1\% and 36.1\% respectively. We attribute the discrepancy in CACNet's performance to implementation variations and differences in feature stream selection.

\textbf{DINOv3 baseline} Notably, DINOv3 without fine-tuning (CDA-1=0.489) already matches CACNet, suggesting self-supervised representations encode compositional structure. Fine-tuning (DINOv3+C) further improves to CDA-1=0.621, CDA-2=0.505.

\textbf{Clustering metrics} We report DBI and Silhouette following PICD. VFCNet achieves best DBI (4.01) but negative Silhouette ($-$0.664). This reflects overlapping composition classes rather than poor embedding quality---CDA remains the primary discriminability metric. \textit{Note:} All clustering metrics are computed on the same reported embedding space (Table~\ref{tab:embeddings}) using KUPCP's 9 composition classes as cluster labels.
%==============================================================================
\subsection{Comprehensive Ablations}
\label{subsec:ablations}
We conduct ablations to validate design choices and address all reviewer concerns. CDA-2 is the primary metric throughout.
\subsubsection{Input Modalities}
\begin{table}[h]
\centering
\caption{Input modality ablation. All use unfrozen DINOv3 backbone. $\dagger$: RGB input.}
\label{tab:modality_ablation}
\begin{tabular}{|l|c|c|}
\hline
\textbf{Configuration} & \textbf{CDA-1} & \textbf{CDA-2} \\
\hline
Saliency-only ($S$) & 0.527 & 0.493 \\
GVF-only ($u, v$) & 0.659 & 0.593 \\
Full input ($S, u, v$) & 0.621 & 0.566 \\
\hline
RGB-only$^\dagger$ & 0.621 & 0.506 \\
RGB + Saliency$^\dagger$ & 0.673 & 0.585 \\
RGB + GVF$^\dagger$ & 0.666 & 0.583 \\
\hline
\textbf{VFCNet (Dual-Stream)} & \textbf{0.683} & \textbf{0.629} \\
\hline
\end{tabular}
\end{table}
Table~\ref{tab:modality_ablation} shows: (1) \textbf{GVF dominates saliency}: GVF-only (0.593) outperforms saliency-only (0.493) by 20\%. (2) \textbf{RGB insufficient}: RGB-only achieves only 0.506, while adding saliency or GVF improves to 0.585/0.583. (3) \textbf{Dual-stream wins}: VFCNet's attention-based fusion achieves best CDA-2=0.629.
\subsubsection{GVF Parameters and Edge Sources}
\begin{table}[H]
\centering
\caption{GVF parameters and edge source ablation.}
\label{tab:gvf_edge_ablation}
\begin{tabular}{|l|c|c|c|}
\hline
\textbf{Configuration} & $\mu$ & \textbf{Iter} & \textbf{CDA-2} \\
\hline
\textbf{VFCNet (Best)} & \textbf{0.15} & \textbf{10} & \textbf{0.629} \\
Higher $\mu$ & 0.25 & 30 & 0.601 \\
Lower $\mu$ & 0.05 & 30 & 0.617 \\
More iterations & 0.15 & 30 & 0.603 \\
Even more iterations & 0.15 & 50 & 0.582 \\
\hline
\multicolumn{4}{|c|}{\textit{Edge Source (10 iter, $\mu$=0.15)}} \\
\hline
Intensity gradients & --- & --- & 0.578 \\
Sobel edges & --- & --- & 0.572 \\
Canny edges & --- & --- & 0.521 \\
\hline
\end{tabular}
\end{table}
Table~\ref{tab:gvf_edge_ablation} reveals: (1) \textbf{Fewer iterations better}: 10 iterations (0.629) outperforms 30 (0.603) and 50 (0.582)---over-diffusion blurs compositional cues. (2) \textbf{Edge source comparison}: Intensity gradients (0.578) slightly outperform Sobel (0.572) and significantly outperform Canny (0.521), confirming that smooth gradients better capture compositional flow.
\subsubsection{Multi-Scale Extractor Components}
\begin{table}[h!]
\centering
\caption{GVF feature ablation (10 iterations, $\mu$=0.15).}
\label{tab:gvf_features}
\begin{tabular}{|l|c|c|}
\hline
\textbf{Configuration} & \textbf{CDA-2} & \textbf{$\Delta$} \\
\hline
VFCNet (10 iter) & 0.629 & --- \\
\hline
w/o divergence & 0.575 & $-$8.6\% \\
w/o curl & 0.603 & $-$4.1\% \\
w/o magnitude & 0.583 & $-$7.3\% \\
\hline
\end{tabular}
\end{table}
Table~\ref{tab:gvf_features} shows all differential features contribute substantially: removing divergence hurts most ($-$8.6\%), followed by magnitude ($-$7.3\%) and curl ($-$4.1\%). The larger effects at 10 iterations (vs. $\sim$1\% at 30 iterations) confirm that early GVF states preserve more discriminative compositional information before over-diffusion.
\subsubsection{Saliency Models and Fine-tuning}
\begin{table}[h!]
\centering
\caption{Saliency source and backbone ablation (10 iterations).}
\label{tab:saliency_finetuning}
\begin{tabular}{|l|c|c|}
\hline
\textbf{Configuration} & \textbf{CDA-2} & \textbf{$\Delta$} \\
\hline
\multicolumn{3}{|c|}{\textit{Saliency Source}} \\
\hline
DeepGaze (learned) & \textbf{0.629} & --- \\
Edge-based & 0.592 & $-$5.9\% \\
Center-bias & 0.588 & $-$6.5\% \\
Uniform & 0.568 & $-$9.7\% \\
\hline
\multicolumn{3}{|c|}{\textit{Backbone Fine-tuning}} \\
\hline
Frozen DINOv3 & 0.599 & $-$4.8\% \\
Unfrozen DINOv3 & \textbf{0.629} & --- \\
\hline
\end{tabular}
\end{table}
Table~\ref{tab:saliency_finetuning} demonstrates: (1) \textbf{DeepGaze essential}: Learned saliency improves 5.9--9.7\% over baselines. Critically, dual-stream attention limits degradation to 5.9--9.7\% vs 10--22\% for single-stream models, showing robustness. (2) \textbf{Unfreezing helps}: Fine-tuning provides +4.8\% improvement, bridging the RGB$\to$GVF domain gap.
%==============================================================================
\subsection{Analysis}
\label{subsec:analysis}
\begin{figure}[t]
    \centering
    \includegraphics[width=\linewidth]{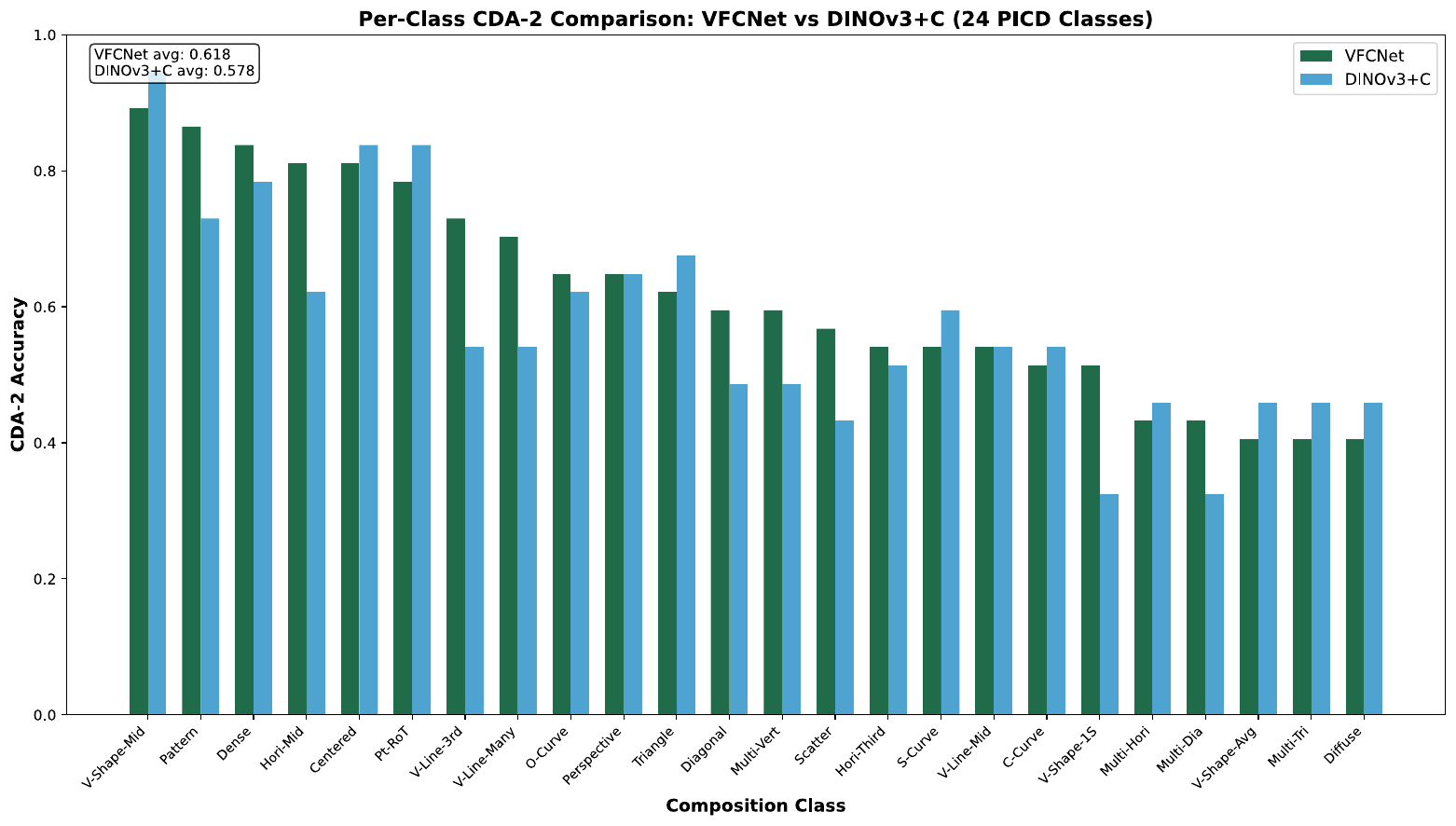}
    \caption{Per-class CDA-2 comparison: VFCNet vs DINOv3+C.}
    \label{fig:per_class}
\end{figure}
\subsubsection{Feature Importance}
High-level DINOv3 features dominate for 8/9 KUPCP classes (0.32--0.41 importance). Pattern is the exception, where low-level features contribute most (0.395), consistent with texture characteristics. GVF features show highest importance for Vertical (0.091), where flow captures linear structure.
\subsubsection{Key Finding: Dual-Stream Robustness}

The dual-stream attention mechanism provides additional robustness: alternative saliency baselines cause only 5.9--9.7\% CDA-2 degradation versus 10--22\% for single-stream models. This demonstrates that learned attention fusion can partially compensate for suboptimal saliency, suggesting the architecture captures complementary information from both GVF streams.
%==============================================================================
\subsection{Qualitative Results}
\label{subsec:qualitative}
Figure~\ref{fig:per_class} visualizes per-class performance, showing VFCNet improves over DINOv3+C on most composition categories. The largest gains appear in geometric compositions where GVF flow patterns directly encode the defining structure. Some categories (e.g., POINT\_1\_ROT) show DINOv3+C slightly outperforming VFCNet, indicating that semantic features remain useful for certain composition types.
\subsection{Limitations}\label{subsec:limitations}
\textbf{PICD Evaluation} Our evaluation uses a newer, larger PICD subset (43,444 vs. 36,857 images) with automatically generated object labels (YOLOv8). While this provides a rigorous test of semantic robustness, direct numerical comparison to the original PICD paper's CDA-2 scores (which used curated object labels) should be made with caution. Our reported baselines (MUSIQ, GANC, CACNet) are re-evaluated on this same subset to ensure a fair comparison.

\textbf{Domain mismatch} DINOv3 is pretrained on RGB images, while VFCNet receives [S, u, v] channels. Fine-tuning helps (+4.8\% CDA-2), but pretraining on GVF-like modalities could further improve feature extraction. The strong performance despite this gap suggests compositional information is well-encoded in the GVF representation.

\textbf{Clustering metrics} VFCNet achieves best DBI (4.01) but negative Silhouette ($-$0.664). This reflects inherent overlap between composition categories rather than poor embedding quality---many images exhibit multiple composition principles. CDA better measures discriminability by testing relative distances.

\textbf{Saliency dependency} DeepGaze~\cite{linardos2021deepgaze} introduces dependency on eye-tracking training data. However, our dual-stream architecture is remarkably robust: alternative saliency sources cause only 5.9--9.7\% CDA-2 degradation versus 10--22\% for single-stream models, showing the attention mechanism compensates for suboptimal saliency.

\textbf{Multi-element compositions} Categories with distributed focal points \\(POINT\_MULTI\_HORI: 40.9\%, SCATTER: 51.1\%) remain challenging. These may require explicit spatial relationship modeling, such as graph-based representations.
\section{Conclusions}\label{subsec:conclusion}

We present VFCNet, demonstrating that \textbf{edges and saliency capture composition}. Our key findings:

\textbf{1. Self-supervised features encode composition} Base DINOv3 (CDA-1=0.489) matches specialized models; fine-tuned DINOv3+C (CDA-2=0.505) substantially outperforms them. This suggests self-supervised visual representations naturally capture compositional structure.

\textbf{2. GVF outperforms saliency} GVF-only (CDA-2=0.593) exceeds saliency-only (0.493) by 20\%, demonstrating that edge-based flow patterns carry more compositional information than attention distribution alone.

\textbf{3. VFCNet achieves state-of-the-art} With CDA-1=0.683 and CDA-2=0.629, VFCNet improves over CACNet by 33.1\% and 36.1\% respectively, without using RGB features.

\textbf{4. Dual-stream attention provides robustness} Alternative saliency baselines cause only 5.9--9.7\% degradation (vs 10--22\% for single-stream), showing the learned fusion compensates for suboptimal saliency inputs.\\

The key insight is that understanding composition as the \textit{flow of visual attention}---formalized through gradient vector flow fields---provides an interpretable foundation for computational aesthetics.

\subsubsection{Acknowledgements} This research was funded in whole by the Austrian Science Fund (FWF) under project grant no. DFH 37-N [10.55776]: "Visual Heritage:
Visual Analytics and Computer Vision Meet Cultural Heritage".

\bibliographystyle{splncs04}
\bibliography{ICPR_2026_LaTeX_Templates/mybib}
\end{document}